\documentclass[runningheads]{llncs}

\usepackage{eccv}

\usepackage{eccvabbrv}

\usepackage{graphicx}
\usepackage{booktabs}
\usepackage{threeparttable}
\usepackage{tabularx}
\usepackage{multirow}
\usepackage{subcaption}
\usepackage{colortbl} 
\usepackage{array}
\definecolor{low}{HTML}{D6EAF8}
\definecolor{high}{HTML}{FADBD8}
\definecolor{grey}{HTML}{EAE7E7}
\newcolumntype{C}{>{\centering\arraybackslash}X} % centered

\usepackage[accsupp]{axessibility}  % Improves PDF readability for those with disabilities.

\usepackage{hyperref}

\makeatletter
\newcommand{\printfnsymbol}[1]{%
  \textsuperscript{\@fnsymbol{#1}}%
}
\makeatother

\usepackage{orcidlink}

\begin{document}

\title{EC-Depth: Exploring the Consistency of Self-supervised Monocular Depth Estimation in Challenging Scenes} 

\titlerunning{EC-Depth}

% \footnote{Equal Conreibution.}

\author{Ziyang Song\thanks{Equal contribution.}\orcidlink{0009-0009-6348-8713} \and
Ruijie Zhu\printfnsymbol{1}\orcidlink{0000-0001-6092-0712} \and
Chuxin Wang\orcidlink{0000-0003-1431-7677} \and
Jiacheng Deng\orcidlink{0000-0003-2838-0378} \and  \\
Jianfeng He\orcidlink{0000-0002-7700-788X} \and
Tianzhu Zhang\orcidlink{0000-0003-0764-6106}
}

\authorrunning{Song et al.}

\institute{University of Science and Technology of China, HeFei, P.R.China\\
% \email{\{songziyang, ruijiezhu, wcx0602, dengjc, hejf\}@mail.ustc.edu.cn;\\
% tzzhang@ustc.edu.cn}
}

\maketitle
% \footnote{*Equal contribution.  $^\dagger$Corresponding author.}

\vspace{-2em}

\begin{abstract}
Self-supervised monocular depth estimation holds significant importance in the fields of autonomous driving and robotics. 
However, existing methods are typically trained and tested on standard datasets, overlooking the impact of various adverse conditions prevalent in real-world applications, such as rainy days. 
As a result, it is commonly observed that these methods struggle to handle these challenging scenarios.
To address this issue, we present EC-Depth, a novel self-supervised two-stage framework to achieve robust depth estimation.
In the first stage, we propose depth consistency regularization to propagate reliable supervision from standard to challenging scenes. 
In the second stage, we adopt the Mean Teacher paradigm and propose a novel consistency-based pseudo-label filtering strategy to improve the quality of pseudo-labels, further improving both the accuracy and robustness of our model.
%
% Extensive experiments substantiate the effectiveness of the proposed method. 
%
Extensive experiments demonstrate that our method achieves accurate and consistent depth predictions in both standard and challenging scenarios, surpassing existing state-of-the-art methods on KITTI, KITTI-C, DrivingStereo, and NuScenes-Night benchmarks.
% , demonstrating its potential for enhancing the reliability of self-supervised monocular depth estimation models in real-world applications.

\end{abstract}

\section{Introduction}
\label{sec:intro}

Depth estimation is a fundamental task in computer vision, with wide-ranging applications in autonomous driving~\cite{schon2021mgnet, wang2019pseudo, you2019pseudo}, scene reconstruction~\cite{yang2020mobile3drecon, niemeyer2022regnerf, li2021hierarchical}, and virtual/augmented reality~\cite{luo2020consistent, noraky2019low}. 
Compared to direct depth acquisition through 3D sensors (e.g. LiDAR), estimating depth from a single image has garnered widespread attention due to its cost-effectiveness and easy deployment. 
Although existing supervised Monocular Depth Estimation (MDE) methods~\cite{bhat2023zoedepth, Ranftl2022, eftekhar2021omnidata, zhu2023ha} can produce accurate depth predictions, they necessitate the gathering of depth annotations, which is both time-consuming and labor-intensive.

\begin{figure}[t]
% \begin{subfigure}{0.63\linewidth}
    \begin{minipage}[t]{0.63\linewidth}
    % \centering
    % \vspace{-0.5em}
    \includegraphics[width=\linewidth]{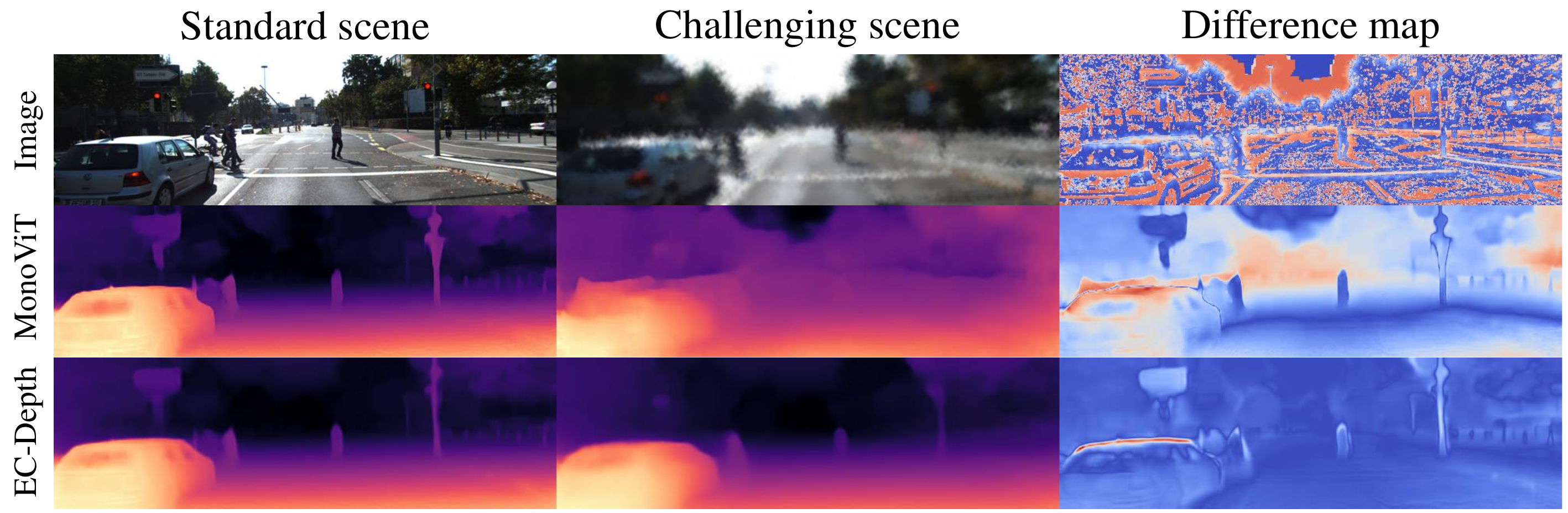}
    \vspace{-1.2em}
    \subcaption{}
    \label{fig:moti}
    \end{minipage}
% \end{subfigure}
% \begin{subfigure}{0.33\linewidth}
    \hspace{0.5em}
    \begin{minipage}[t]{0.33\linewidth}
    \centering
    % \vspace{-0.5em}
    \includegraphics[width=\linewidth]{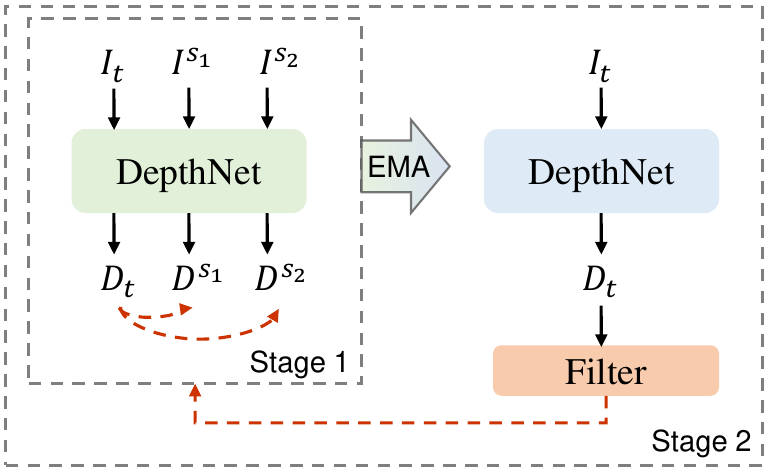}
    \vspace{-1.2em}
    % \caption{\textbf{The illustration of EC-Depth.} We adopt consistency regularization and the Mean Teacher paradigm for reliable supervision in challenging scenes.} 
    \subcaption{}
    \label{fig:framework_concise}
    \end{minipage}
% \end{subfigure}
\vspace{-0.5em}
\caption{\textbf{(a) Comparison of depth predictions in standard and challenging scenes.}
The third column shows differences in input images and depth predictions, with blue indicating consistent region. 
\textbf{(b) The overall structure of EC-Depth.} We adopt consistency regularization and the Mean Teacher paradigm for reliable supervision in challenging scenes, significantly enhancing the robustness of our model.
} 
\vspace{-1.5em}
\end{figure}

To address the above issue, many self-supervised monocular depth estimation methods~\cite{zhou2017unsupervised, godard2017unsupervised,godard2019digging,pillai2019superdepth,almalioglu2019ganvo,guizilini20203d,zhao2022monovit} have emerged. 
Based on the rigid scene assumption, \ie, all objects in the scene are static, these methods leverage the geometric consistency between consecutive frames for depth supervision. 
Specifically, they train a PoseNet and a DepthNet to generate camera ego-motion and depth prediction, which are then utilized to synthesize the current frame from neighboring frames.
By constraining the photometric consistency between the synthesized image and the real image, the model is guided to learn depth predictions in a self-supervised way.
Based on this paradigm, existing self-supervised MDE methods~\cite{zhao2022monovit,peng2021excavating,wang2023planedepth} demonstrate satisfactory performance in standard outdoor scenes.
However, when encountering challenging scenarios such as rainy and snowy days, they exhibit significant accuracy degradation (see row 2 in~\Cref{fig:moti}).
After analyzing the failure cases of depth estimation in challenging scenes, we find that the presence of significant moving noise (raindrops, motion blur, \etc.) in these scenarios violates the rigid scene assumption, resulting in unreliable supervision.
Recent studies~\cite{wang2021regularizing,saunders2023self,gasperini2023robust} explore to enhance the robustness of models in such challenging and harsh scenarios. However, they still rely on the photometric loss for supervision in noisy and dynamic scenarios, which is not always reliable.

In semi-supervised approaches, consistency regularization~\cite{sohn2020fixmatch, yang2023revisiting, hendrycks2019augmix} and Mean Teacher~\cite{tarvainen2017mean} are two effective ways to provide reliable supervision for unlabeled data. 
We argue that such techniques can also be applied for depth estimation in challenging scenarios based on the following reasons:
(1) In the same scene, images under challenging conditions should share consistent depth with that under standard condition, and models tend to predict more accurate depth in standard scenes.
Therefore, we can simulate challenging conditions on standard images, so that consistency regularization can be applied to propagate supervision from standard scenes to challenging scenarios.
(2) 
Mean Teacher can provide stable pseudo-labels and boost the performance of the student model by performing exponentially moving average (EMA) integration of historical knowledge. 
However, not all the generated pseudo-labels are trust-worthy in self-supervised depth estimation. 
Therefore, we argue that filtering out unreliable depth predictions can improve the overall quality of pseudo-labels, thus benefiting the model performance in challenging scenarios.

Based on the above discussion, we propose a novel two-stage framework named EC-Depth to explore the consistency of self-supervised monocular depth estimation in challenging scenarios (see~\Cref{fig:framework_concise}).
\textbf{In the first stage}, we simulate images in challenging scenarios from standard scenes and create an image triplet consisting of the origin image and two simulated challenging images. 
For the standard image, we apply the photometric loss as depth supervision.
For challenging images, we constrain the consistency of their depth predictions with that of the standard image, thereby propagating reliable supervision from standard to challenging scenes.
The depth consistency regularization enables the model to maintain high accuracy on standard benchmarks while realizing robust and consistent depth predictions in challenging scenarios.
\textbf{In the second stage}, we further distill the model of the first stage using the Mean Teacher paradigm. 
Specifically, we use the teacher network to generate pseudo-labels as the supervision of the student network.
To improve the quality of pseudo-labels, a consistency-based pseudo-label filtering strategy is introduced, which selects reliable depth predictions that simultaneously satisfy geometric consistency between consecutive frames and depth consistency under different perturbations.
Compared with previous methods, our approach exhibits significant advantages in terms of accuracy and consistency in depth predictions across standard and challenging scenes (see row 3 in~\Cref{fig:moti}).

In summary, the main contributions of our work are as follows:
\vspace{-0.6em}
\begin{enumerate}
    \item[--] We introduce a novel two-stage self-supervised depth estimation framework named EC-Depth, which can improve the accuracy and robustness of the depth estimation model specifically in challenging scenarios.
    \item[--] To generate effective supervision for challenging scenes, we simulate various perturbations and design a perturbation-invariant depth consistency constraint to propagate supervision from standard to challenging scenes.
    \item[--] We introduce Mean Teacher to distill the model and devise a consistency-based pseudo-label filtering strategy to ensure reliable depth supervision.
    \item[--] The proposed method significantly outperforms other methods on the challenging KITTI-C benchmark while maintaining accuracy on KITTI dataset. Furthermore, our model exhibits an exceptional generalization capability in zero-shot tests on DrivingStereo and NuScenes-Night datasets.
\end{enumerate}

\section{RELATED WORKS}

\subsection{Self-supervised Monocular Depth Estimation}

Self-supervised monocular depth estimation has gained significant attention due to its ability to train models without the need for depth annotations. 
Existing methods can be categorized into two types:
one type~\cite{garg2016unsupervised, godard2017unsupervised} utilizes geometric constraints from stereo image pairs to learn depth, while the other~\cite{zhou2017unsupervised, godard2019digging} relies on geometric constraints from consecutive video frames. 
In essence, they both generate self-supervised signals through viewpoint synthesis.
SfMLearner~\cite{zhou2017unsupervised} stands as a pioneering work in employing view synthesis techniques. %
Following its procedure, self-supervised methods simultaneously train a depth estimation network and a pose estimation network to synthesize the current frame using neighboring frames or image pairs. 
And the model is trained by enforcing photometric consistency between the current frame and the synthesized images.
Subsequent advances have been achieved in several aspects, including improving network architecture~\cite{zhou_diffnet, yan2021channel, zhao2022monovit}, designing optimization strategies and loss functions~\cite{shu2020feature}, exploiting multi-frame information~\cite{watson2021temporal} and plane information~\cite{wang2023planedepth}, uncertainty modeling~\cite{poggi2020uncertainty}, and adopting pseudo-labels~\cite{Petrovai_2022_CVPR, ren2022adaptive, zhou2021sub}.
However, these methods only consider depth estimation in standard scenes and often performs poorly in challenging scenarios such as rainy or foggy weather. 

\subsection{Robust Monocular Depth Estimation}
Ensuring robust performance in challenging scenarios holds paramount importance, particularly in safety-critical applications like autonomous driving. 
A significant portion of the images in challenging scenarios is occupied by moving noise, which causes the model to deviate from the rigid scene assumption. 
Therefore, directly applying photometric loss to challenging scenario images is not feasible.
Several prior studies have endeavored to address this issue.
Thermal imaging camera~\cite{kim2018multispectral, lu2021alternative} is one of the common sensors used to alleviate the challenges caused by low visibility, offering improved adaptability to nighttime scenes.
However, the images captured have limited texture details and low resolution.
Therefore, some methods~\cite{wang2021regularizing, guo2020zero, jiang2021enlightengan} begin to explore denoising techniques for challenging images. Typically, they exploit networks to enhance the brightness or eliminate noise, making the images appear close to standard scene images.
Considering the additional overhead caused by denoising, some other methods exploit a simpler approach, namely, introducing noise.
Robust-Depth~\cite{saunders2023self} and MD4all~\cite{gasperini2023robust} simulate images in challenge scenes from a standard dataset and modify the photometric loss to supervise depth predictions for challenging scenes.
Different from them, we leverage consistency regularization to generate supervision for challenging scenes.
The ablation studies demonstrate that the proposed consistency regularization provides more direct guidance for depth estimation in challenging scenes.
Besides, the proposed self-distillation further boosts the performance of our model in both standard and challenging scenarios.

\subsection{Semi-supervised Learning} 
The core issue in semi-supervised learning is how to design reasonable and effective supervision for unlabelled data.
Consistency regularization~\cite{sohn2020fixmatch, zhang2021flexmatch, xie2020unsupervised, hendrycks2019augmix, yang2023revisiting} and Mean Teacher~\cite{tarvainen2017mean, cao2024hassod, xu2021end, zheng2021exploiting} are two leading solutions to tackle the problem.
Consistency regularization assumes that predictions of an unlabeled example should be invariant to different forms of perturbations.
The Mean Teacher paradigm aims to enhance the generalization performance of deep learning models on unlabeled data. The teacher maintains a snapshot of the historical students using EMA strategy, providing more stable and experienced pseudo-labels.
Both solutions help the model mitigate the sensitivity to noise and outliers in the training data.
Inspired by these methods, we innovatively introduce their ideas to the self-supervised depth estimation task.
As a result, our model not only achieves state-of-the-art performance in challenging scenarios but also demonstrates exceptional zero-shot generalization performance across multiple datasets.

\section{METHOD}
\subsection{Preliminary}

Given a single frame $I_t$, monocular depth estimation aims to predict its corresponding depth map $D_t$.
In a self-supervised setting, model supervision comes from adjacent frames $I_{{t}^{\prime}} \in \left \{ I_{t-1},I_{t+1} \right \}$.
Specifically, following the paradigm of SFMLearner~\cite{zhou2017unsupervised} and Monodepth2~\cite{godard2019digging}, self-supervised monocular depth estimation methods simultaneously train a depth estimation network to predict the depth of the current frame $D_t=\mbox{DepthNet}(I_t)$ and a pose estimation network to estimate the camera ego-motion to next timestamp $T_{t \to t^{\prime}}=\mbox{PoseNet}(I_t, I_{t^{\prime}})$. 
Then, $I_{{t}^{\prime}}$ can be projected onto the current timestamp to generate a synthesized counterpart:
\begin{equation} \label{eq:basewarp}
    I_{t^{\prime} \to t} = I_{t^{\prime}}\left \langle \mbox{proj}\left ( D_t,T_{t \to t^{\prime}},K \right )  \right \rangle ,
\end{equation}
where $K$ is the camera intrinsics, $\mbox{proj}(\cdot)$ operator returns the 2D coordinates of $D_t$ when reprojected into the camera of $I_{t^{\prime}}$, and $\langle \cdot \rangle$ is the pixel sampling operator. 
If the prediction is accurate, $I_{t^{\prime} \to t}$ is supposed to be identical to $I_t$.
So we can enforce the photometric consistency between these two images to provide effective supervision.
The photometric loss is formulated as  
\begin{equation}
    L_p = \min_{t^{\prime} }\mbox{pe}(I_t,I_{t^{\prime} \to t}) ,
\end{equation}
\vspace{-0.5em}
\begin{equation}
    \mbox{pe}\left ( I_a,I_b \right ) = \frac{\theta}{2} \left(1-\mbox{SSIM}\left(I_a,I_b\right)\right) + \left ( 1-\theta \right ) \left \| I_a-I_b \right \|_1 ,
\end{equation}
where $\mbox{pe}(\cdot)$ is a combination of SSIM~\cite{wang2004image} and $L_1$ loss that measures the difference between two images.
Additionally, the edge-aware smoothness loss $L_{e}$~\cite{Ranjan_2019_CVPR} is widely adopted to deal with depth discontinuities:
\begin{equation}
    L_{e}(D)=|\partial_xw(D)|e^{\partial_xI}+|\partial_yw(D)|e^{\partial_yI},
\end{equation}
where $w(D)$ is the normalized inverse depth of $D$, and $\partial_x$ and $\partial_y$ are the horizontal and vertical gradients, respectively.

\begin{figure*} [t]
\centering
\includegraphics[width=\textwidth]{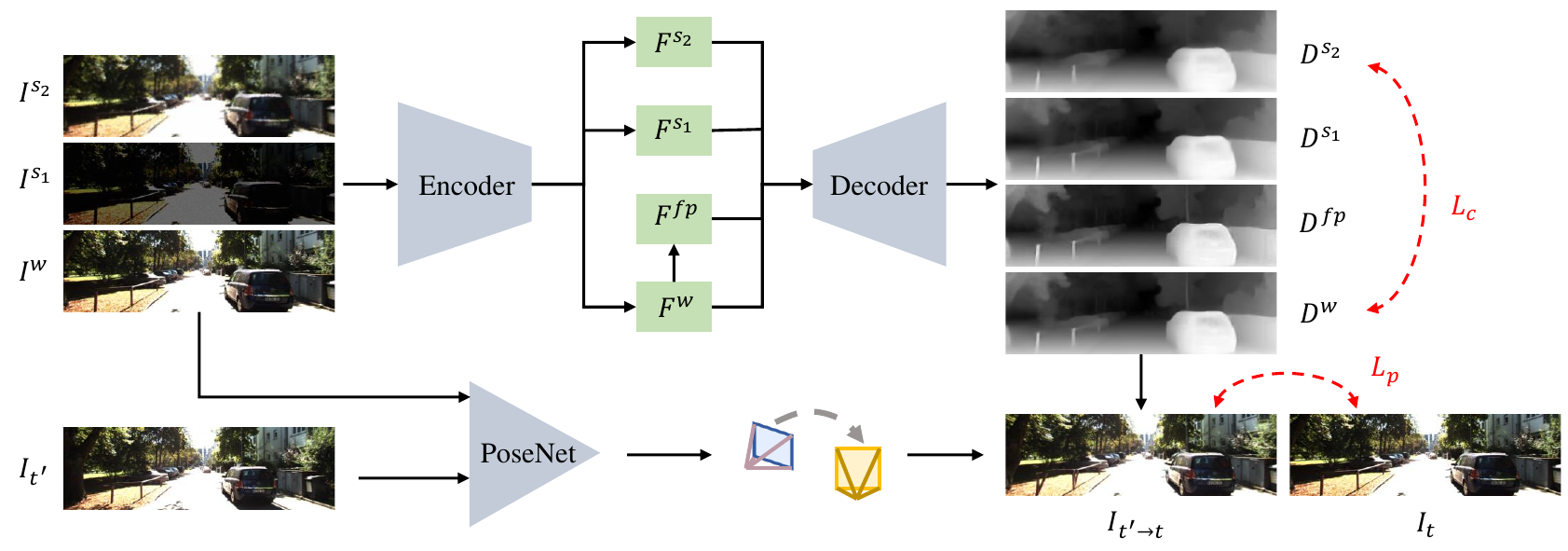}
\vspace{-2em}
\caption{\textbf{The first-stage training framework of EC-Depth. }
In the first stage, we train DepthNet and PoseNet with the perturbation-invariant depth consistency loss. } 
\label{fig:framework_stage1}
\vspace{-1em}
\end{figure*}

\subsection{Overall achitecture}
\begin{figure*} [t]
\centering
\includegraphics[width=\textwidth]{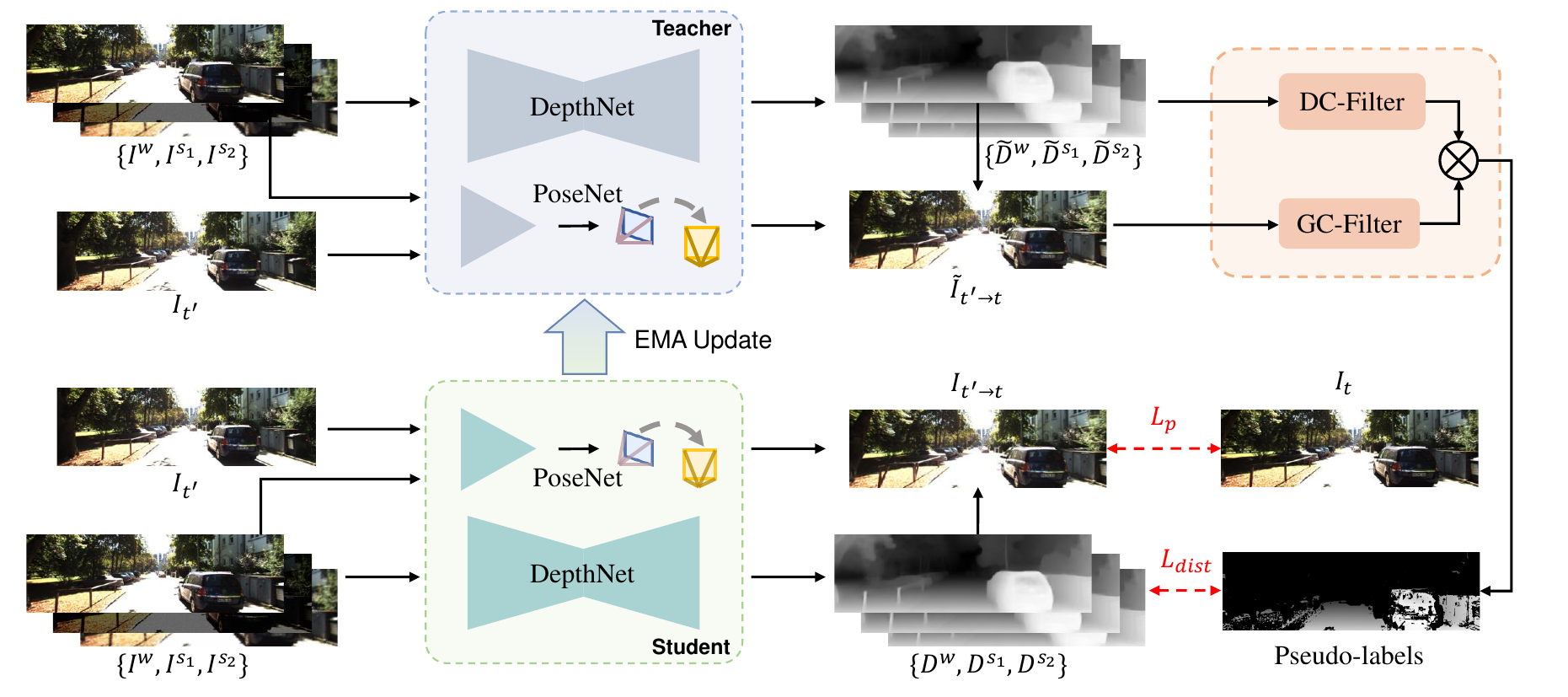}
\caption{
\textbf{The second-stage training framework of EC-Depth.} 
In the second stage, we leverage the Mean Teacher paradigm to generate pseudo-labels for self-distillation. In particular, we propose a depth consistency-based filter (DC-Filter) and a geometric consistency-based filter (GC-Filter) to filter out unreliable pseudo-labels.} \label{fig:framework_stage2}
\vspace{-1em}
\end{figure*}

The frameworks of our two training stages are illustrated in Fig.~\ref{fig:framework_stage1} and Fig.~\ref{fig:framework_stage2}, respectively.
Our method is agnostic to the design of depth networks, which allows it to be easily transferred to any self-supervised monocular depth estimation method.
In this paper, we take MonoViT~\cite{zhao2022monovit} as our baseline.
In the first stage, we introduce weak-to-strong perturbations to construct a diverse perturbation space, operating at both the image and feature levels. 
For depth predictions of weakly perturbed images, we adopt the photometric loss following previous works~\cite{godard2019digging}.
To effectively supervise depth predictions on strongly perturbed images, we devise a perturbation-invariant depth consistency loss. 
In the second stage, we employ the Mean Teacher paradigm to further distill the model of the first stage with pseudo-labels. 
To select accurate and robust depth pseudo-labels, we design a consistency-based pseudo-label filtering strategy based on geometric consistency between consecutive frames and depth consistency under different perturbations, respectively.
The teacher iteratively integrates information from historical students using exponential moving average (EMA), yielding more reliable and stable depth pseudo-labels.

\subsection{Perturbation-invariant depth consistency regularization}
% \subsection{Weak-to-strong perturbation stream}
Consistency Regularization is a powerful technique in semi-supervised learning, which is widely applied across various tasks~\cite{hu2021semi,melas2021pixmatch,xu2022cross}.
The core idea of this approach is to encourage the model to produce consistent outputs under different perturbations, thereby enhancing the generalization of the model. 
Motivated by this, we apply this concept to self-supervised monocular depth estimation tasks, aiming to enhance the robustness of the model in challenging scenarios. 
Due to the difficulty and inaccuracy in collecting depth annotations in challenging scenarios~\cite{gasperini2023robust}, we propose weak-to-strong image perturbations to construct the perturbation space.
Furthermore, we design a perturbation-invariant depth consistency loss to encourage consistent depth predictions under different perturbations, thus resulting in accurate depth estimation in challenging scenarios.

\textbf{Weak-to-strong image perturbations. }
Given an image $I_t$ in the standard scene, we use simple color jitter to obtain a weakly perturbed image $I^w$:
\begin{equation}
    I^w = \mbox{color}(I_t) 
\end{equation}
where $\mbox{color}(\cdot)$ operator slightly changes the brightness, contrast, saturation, and hue of the image.
Then, we construct strong perturbations by using graphical transformations to obtain a strongly perturbed image $I^s$:
\begin{equation}
    I^s = \mbox{corrupt}(I_t) 
    \label{eq:strong}
\end{equation}
where $\mbox{corrupt}(\cdot)$ operator is randomly sampled from 18 types of image perturbations~\cite{hendrycks2019benchmarking}, including different weather or light conditions, sensor failures or movement, and the noises during data processing. 
Within a mini-batch, we sample one weakly perturbed image $I^w$ and two strongly perturbed images $I^{s_1}, I^{s_2}$ to form an image triplet, which is then sent into $\mbox{DepthNet}$ to obtain their corresponding depth predictions $D^w, D^{s_1}, D^{s_2}$.
The image-level perturbations inject prior heuristic knowledge into the model, helping it achieve superior performance when confronted with these perturbations. 
However, when encountering unknown perturbation, the model's performance still diminishes~\cite{yang2023revisiting}. 
Therefore, we explore a broader perturbation space through feature-level perturbation, which is accomplished with a simple channel dropout: 
\begin{equation}
    F^{fp} = \mbox{Dropout}(F^w),
\end{equation}
where $F^w$ is the feature of $I^w$ and $ F^{fp} $ is the feature after feature-level perturbation. 
The perturbed feature is then fed into the depth decoder to obtain its according depth prediction $D^{fp}$.
Through this technique, our model is equipped with better adaptation to unseen perturbations.

\textbf{Perturbation-invariant depth consistency loss.}
To effectively supervise the depth predictions under different perturbations, we inherit the spirit of consistency regularization and design a novel perturbation-invariant depth consistency loss as: 
\begin{equation}
    L_{c}=\frac{1}{K}\sum_{i}^{} w\left(D^i\right ) \log{\frac{w\left(D^i\right )}{\frac{1}{K}\sum_{i}{w\left(D^i\right )}}},
\end{equation}
where $i\in \{s_1, s_2, w, fp\}$ , $D^i$ is the depth prediction under the $i$-th different perturbations, $K$ is the number of perturbations within a mini-batch, and $w\left(D_i\right )$ denotes the normalized inverse depth of $D^i$.
Similar in form to the Jensen-Shannon Divergence~\cite{menendez1997jensen}, this loss measures the distance of all depth predictions to their mean values, with smaller distance resulting in smaller loss value. 
When depth predictions under different perturbations are completely consistent, the loss value reduces to 0.
Additionally, we believe that the accuracy of the depth prediction in close regions is more crucial than that in distant regions, especially in applications like autonomous driving.
Therefore, we penalize the depth inconsistencies near the camera with higher weights.
Through the proposed depth consistency loss, we effectively transfer supervision from standard images to challenging images, thereby enhancing the robustness of the model.

\textbf{The first-stage training.}
For the first stage training, we feed the perturbed images $I^w$, $I^{s_1}$, $I^{s_2}$ as a mini-batch to the model and obtain their corresponding depth predictions $D^w$, $D^{s_1}$, $D^{s_2}$ and $D^{fp}$ corresponding to the feature-level perturbation.
Intuitively, the depth prediction $D^w$ is more stable and easy to optimize. 
Therefore, we opt to impose the photometric loss $L_p$ and edge-aware smoothness loss $L_e$ on $D^w$, and impose perturbation-invariant depth consistency loss on all the depth predictions to transfer the knowledge from the weakly perturbed image to strongly perturbed images. 
Finally, we sum up the three losses at each image scale $s$ as the total loss:
\begin{equation}
    L_{stage1}=\frac{1}{N}\sum_{s=1}^{N}\left ( L_p + \alpha L_e + \beta L_c \right ),
\end{equation}
where $s$ indicates scale, $ \alpha, \beta, N$ are set to $0.001, 0.001, 4$ respectively. 

\subsection{Consistency-based self-distillation}
\label{Pseudo}

To further improve the quality of supervision for challenging scenarios, we exploit the self-distillation framework to distill the model of the firststage. 
Specifically, we initialize the teacher and the student network with the model weights trained in the first stage. 
The depth predictions of the teacher network are then utilized as pseudo-labels to supervise the training of the student. 
During training, the teacher network is updated by the EMA weights of the student network. 
Through weight averaging, the teacher network can integrate historical students' knowledge, resulting in more reliable and stable depth pseudo-labels. 
On the other hand, depth pseudo-labels may not be accurate in some challenging regions, such as complex textured areas.
Therefore, we design a pseudo-label consistency filtering strategy to remove unreliable pseudo-labels in these areas.

\textbf{Consistency-based pseudo-label filtering.}
To provide high quality pseudo-labels, we propose a consistency-based pseudo-label filtering strategy, considering both accuracy and robustness of pseudo-labels.
Firstly, we exploit GC-Filter to filter out pixels that do not satisfy geometric consistency, \ie, inaccurate pixels.
Given the current frame $ I_t $ and its adjacent frame $ I_{t^{\prime}} $, the teacher network predicts the depth map of the current frame $ \tilde{D}^w $ and camera ego-motion $ \tilde{T}_{t \to t^{\prime}} $. Then we can synthesize the current frame $ \tilde{I}_{t^{\prime} \to t} $ and compute the photometric error.
The geometric consistency mask in GC-Filter is then defined as below:
\begin{equation}
    M_{g} = \left[ \min_{t^{\prime} }\mbox{pe}(I_t,\tilde{I}_{t^{\prime} \to t}) < \delta_{g} \right],
\end{equation}
where $ \left[ \cdot \right] $ are the Iverson brackets and $ \delta_{g} $ is a predefined threshold on the reprojection loss. 
Secondly, we use DC-Filter to filter out pseudo-labels which are sensitive to perturbations.
After predicting the corresponding depth predictions $ \{ \tilde{D}^w, \tilde{D}^{s_1} , \tilde{D}^{s_2} \} $ of the image triplet by the teacher network, we define the depth consistency mask in DC-Filter as below:
\begin{equation}
    M_{d} = \left[ \left| \tilde{D}^{s_1} - \tilde{D}^w \right| < \delta_{d} \right] \odot \left[ \left| \tilde{D}^{s_2} - \tilde{D}^w \right| < \delta_{d} \right],
\end{equation}
where $ \odot $ is the pixel-wise product and $ \delta_{d} $ is a predefined threshold for the absolute depth difference values. 
By combining these two filters, we successfully select pseudo-labels that satisfy both geometric and depth consistency, providing high-quality pseudo-labels for model distillation.

\textbf{The second-stage training.} 
Integrating both two consistency-based filters, we select reliable pseudo-labels and define a distillation loss to supervise the student network:
\begin{equation}
    L_{dist} =  \frac{1}{K}\sum_{i}^{} \left| \tilde{D}^w - D^i \right| \odot M_{g} \odot M_{d},
\end{equation}
where $i\in \{s_1, s_2, w, fp\}$ , $D^i$ is the depth predictions of the student network under the $i$-th different perturbations.
Finally, we train the student network with the total loss in the second stage as:
\begin{equation}
    L_{stage2}=\frac{1}{N}\sum_{s=1}^{N}\left ( L_p + \alpha L_e + \gamma L_{dist} \right ),
\end{equation}
where above $ \alpha, \gamma $ are set to $0.001, 1$ respectively. 
During inference, we only use DepthNet of the student model to generate depth map for each frame.

\section{Experiments}
\subsection{Implementation Details}
\noindent \textbf{Training.}
The proposed EC-Depth is implemented using Pytorch. We train the model on Nvidia RTX 3090 GPUs with batch size 8 and  
optimize it with the AdamW~\cite{loshchilov2017decoupled} optimizer for 20 epochs.
The learning rate of PoseNet and depth decoder is initially set as 1e-4, while the initial learning rate of pretrained depth encoder is set as 5e-5. 
All the learning rates decay by a factor of 10 at the final 5 epochs.
The experiments are trained with an input resolution of $640 \times 192$.
We set the hyperparameters $\delta_g$, $\delta_d$ to 0.04, 0.04, respectively. 

\noindent \textbf{Evaluation.}
For evaluation metrics, we adopt the widely used seven metrics~\cite{eigen2014depth}: absolute relative difference (AbsRel), square relative difference (SqRel), root mean squared error (RMSE), its log variant (RMSL), and accuracy rates $a_1$, $a_2$, $a_3$. 
The first four metrics assess the error of depth predictions from different perspectives, with smaller values indicating better performance. 
The last three metrics measure the percentage of inlier pixels for three thresholds (1.25, $\rm 1.25^2$, $\rm 1.25^3$), with larger values indicating better performance.

\subsection{Datasets}
\noindent\textbf{KITTI}~\cite{Geiger2013IJRR} is a widely used outdoor benchmark for depth estimation, which contains sequential stereo images and sparse points collected by sensors mounted on vehicles. 
During training, we follow Zhou split~\cite{zhou2017unsupervised}, which contains 19905 training images and 2212 validation images. During evaluation, the Eigen split~\cite{eigen2014depth} with 697 test images is adopted. Besides, the capturing range is set to 0-80m.

\noindent\textbf{KITTI-C}~\cite{kong2024robodepth} is a comprehensive benchmark to evaluate the robustness of monocular depth estimation. 
The benchmark shares the same raw images with the test set of KITTI, but simulates diverse challenging scenarios, including bad weather or lighting conditions, sensor failure or movement, and noises in data processing. 
In total, the benchmark contains 18 types of perturbations with 5 levels of severity.
Following the RoboDepth Challenge~\cite{Kong2023TheRC},
we average the test results across all kinds of perturbations with all levels of severity as the final metrics to compare with the state-of-the-art methods.

\noindent\textbf{DrivingStereo}~\cite{yang2019drivingstereo} is a large-scale real-world autonomous driving dataset. It provides a challenging subset of images under four weather conditions (foggy, cloudy, rainy and sunny), each of which contains 500 images. We test on this subset to evaluate the robustness and generalization of MDE models.

\noindent\textbf{NuScenes}~\cite{caesar2020nuscenes} is a comprehensive autonomous driving dataset comprising 1000 video clips. We select the night-time test split~\cite{wang2021regularizing} to test the robustness of our method in night-time scenarios. The scenes are pretty challenging due to low visibility and complicated traffic conditions.

\begin{table*}[t]\footnotesize
\caption{\textbf{Quantitative results on KITTI and KITTI-C.} 
EC-Depth* is the model of the first stage and EC-Depth is the model of the second stage. For \colorbox{low}{error-based metrics}, lower is better; and for \colorbox{high}{accuracy-based metrics}, higher is better. The best and second best results are marked in \textbf{bold} and \underline{underline}.}\label{table:results}  
\resizebox{\linewidth}{!}{
% \begin{tabular}{c|c|c|c|c|c|c|c|c|c}
\begin{tabularx}{1.2\linewidth}{c|C|*{4}{C}|*{3}{C}}
\hline
Method                                               & Test                         & \cellcolor{low}AbsRel & \cellcolor{low}SqRel & \cellcolor{low}RMSE & \multicolumn{1}{c|}{\cellcolor{low}RMSL} & \cellcolor{high}$ a_1 $ & \cellcolor{high}$ a_2 $ & \cellcolor{high}$ a_3 $             \\ \hline
                               &                     KITTI-C                       & 0.204                                  & 1.871                                  & 6.918                                  & 0.295                                  & 0.692                                  & 0.872                                  & 0.943                                  \\
\multirow{-2}{*}{MonoDepth2\cite{godard2019digging}}   & \cellcolor{grey}KITTI & \cellcolor{grey}0.115          & \cellcolor{grey}0.903          & \cellcolor{grey}4.863          & \cellcolor{grey}0.193          & \cellcolor{grey}0.877          & \cellcolor{grey}0.959          & \cellcolor{grey}0.981          \\
\hline
                               &       KITTI-C                       & 0.205                                  & 1.738                                  & 6.865                                  & 0.295                                  & 0.679                                  & 0.871                                  & 0.944                                  \\
\multirow{-2}{*}{HR-Depth\cite{lyu2021hr}}     & \cellcolor{grey}KITTI & \cellcolor{grey}0.109          & \cellcolor{grey}0.792          & \cellcolor{grey}4.632          & \cellcolor{grey}0.185          & \cellcolor{grey}0.884          & \cellcolor{grey}0.962          & \cellcolor{grey}0.983          \\
\hline
                               &             KITTI-C                       & 0.225                                  & 2.062                                  & 7.152                                  & 0.316                                  & 0.654                                  & 0.846                                  & 0.931                                  \\
\multirow{-2}{*}{CADepth\cite{yan2021channel}}      & \cellcolor{grey}KITTI & \cellcolor{grey}0.107          & \cellcolor{grey}0.803          & \cellcolor{grey}4.592          & \cellcolor{grey}0.183          & \cellcolor{grey}0.890          & \cellcolor{grey}0.963          & \cellcolor{grey}0.983          \\
\hline
                               &               KITTI-C                       & 0.188                                  & 1.622                                  & 6.541                                  & 0.280                                  & 0.722                                  & 0.886                                  & 0.946                                  \\
\multirow{-2}{*}{DIFFNet\cite{zhou_diffnet}}      & \cellcolor{grey}KITTI & \cellcolor{grey}0.102          & \cellcolor{grey}0.753          & \cellcolor{grey}4.459          & \cellcolor{grey}0.179          & \cellcolor{grey}0.897          & \cellcolor{grey}0.965          & \cellcolor{grey}0.983          \\
\hline
                               &            KITTI-C                       & 0.161                                  & 1.292                                  & 6.029                                  & 0.247                                  & 0.768                                  & 0.915                                  & 0.964                                  \\
\multirow{-2}{*}{MonoViT\cite{zhao2022monovit}}      & \cellcolor{grey}KITTI & \cellcolor{grey}\textbf{0.099} & \cellcolor{grey}{\underline{0.708}}    & \cellcolor{grey}4.372          & \cellcolor{grey}{\underline{0.175}}    & \cellcolor{grey}\textbf{0.900} & \cellcolor{grey}\textbf{0.967} & \cellcolor{grey}{\underline{0.984}}    \\
\hline
                               &            KITTI-C                       & 0.185                                  & 1.537                                  & 6.624                                  & 0.277                                  & 0.716                                  & 0.892                                  & 0.952                                  \\
\multirow{-2}{*}{LiteMono\cite{zhou_diffnet}}      & \cellcolor{grey}KITTI & \cellcolor{grey}0.107          & \cellcolor{grey}0.766          & \cellcolor{grey}4.560          & \cellcolor{grey}0.183          & \cellcolor{grey}0.866          & \cellcolor{grey}0.963          & \cellcolor{grey}0.983          \\
\hline

                               &          KITTI-C                       & 0.123                                  & 0.957                                  & 5.093                                  & 0.202                                  & 0.851                                  & 0.951                                  & 0.979                                  \\

\multirow{-2}{*}{Robust-Depth\cite{saunders2023self}} & \cellcolor{grey}KITTI & \cellcolor{grey}{\underline{ 0.100}}    & \cellcolor{grey}0.747          & \cellcolor{grey}4.455          & \cellcolor{grey}0.177          & \cellcolor{grey}0.895          & \cellcolor{grey}{\underline{ 0.966}}    & \cellcolor{grey}{\underline{ 0.984}}    \\ \hline
\specialrule{-1em}{0.2pt}{0.2pt} \\
\hline

                               &              KITTI-C                       & {\underline{ 0.115}}                            & {\underline{ 0.841}}                            & {\underline{ 4.749}}                            & {\underline{ 0.189}}                            & {\underline{ 0.869}}                           & {\underline{ 0.958}}                            & {\underline{ 0.982}}                            \\
\multirow{-2}{*}{EC-Depth*}     & \cellcolor{grey}KITTI & \cellcolor{grey}{\underline{ 0.100}}    & \cellcolor{grey}{\underline{ 0.708}}    & \cellcolor{grey}{\underline{ 4.367}}    & \cellcolor{grey}{\underline{ 0.175}}    & \cellcolor{grey}{\underline{ 0.896}}    & \cellcolor{grey}{\underline{ 0.966}}    & \cellcolor{grey}{\underline{ 0.984}}    \\
\hline
                               &                 KITTI-C                       & \textbf{0.111}                         & \textbf{0.807}                         & \textbf{4.651}                         & \textbf{0.185}                         & \textbf{0.874}                         & \textbf{0.960}                         & \textbf{0.983}                         \\
\multirow{-2}{*}{EC-Depth}    & \cellcolor{grey}KITTI & \cellcolor{grey}{\underline{ 0.100}}    & \cellcolor{grey}\textbf{0.689} & \cellcolor{grey}\textbf{4.315} & \cellcolor{grey}\textbf{0.173} & \cellcolor{grey}{\underline{ 0.896}}    & \cellcolor{grey}\textbf{0.967} & \cellcolor{grey}\textbf{0.985} \\ \hline
\end{tabularx}
}
\vspace{-1em}
\end{table*}

\begin{figure*}[ht] 
\centering
\includegraphics[width=\textwidth]{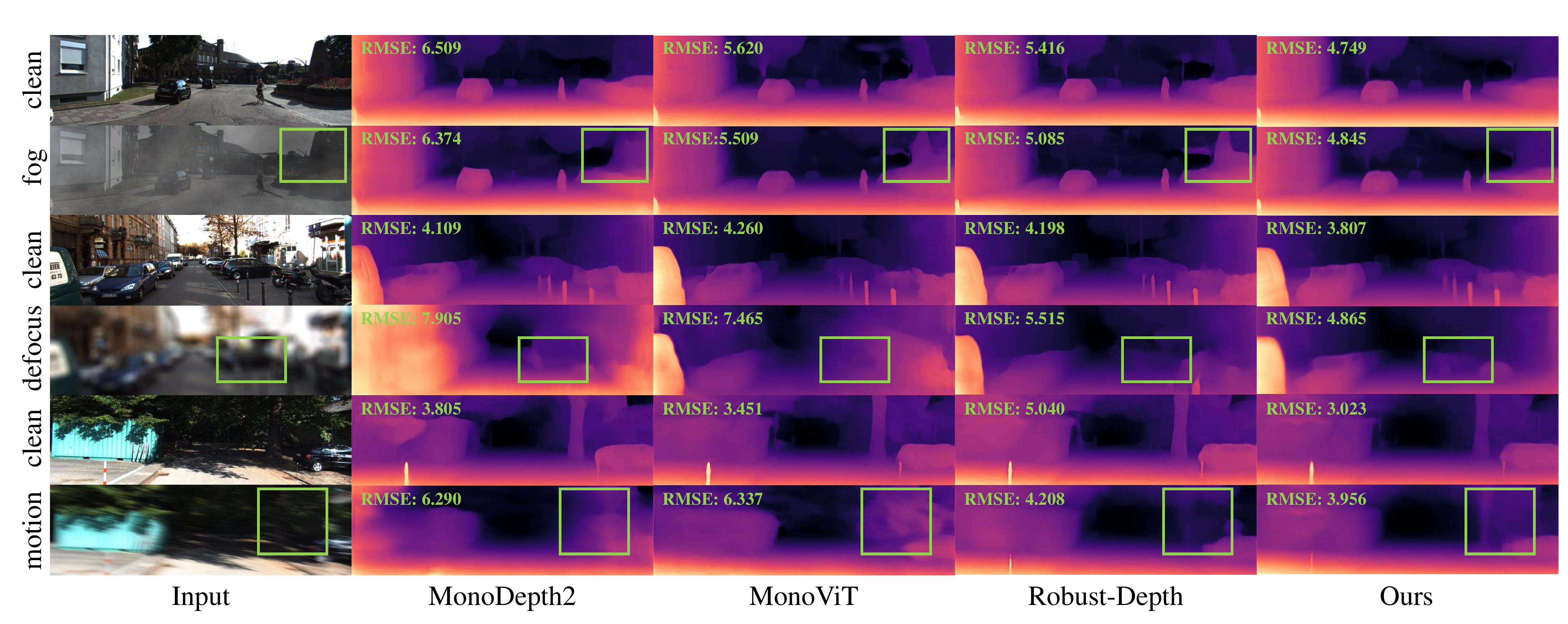}
\vspace{-2em}
\caption{\textbf{Qualitive results on KITTI and KITTI-C benchmark.} Our method can predict accurate and consistent depth maps even under severe perturbations.
}
\label{fig:kittic}
\vspace{-1em}
\end{figure*}

\begin{figure*}[t]
    \centering
    \includegraphics[width=\linewidth]{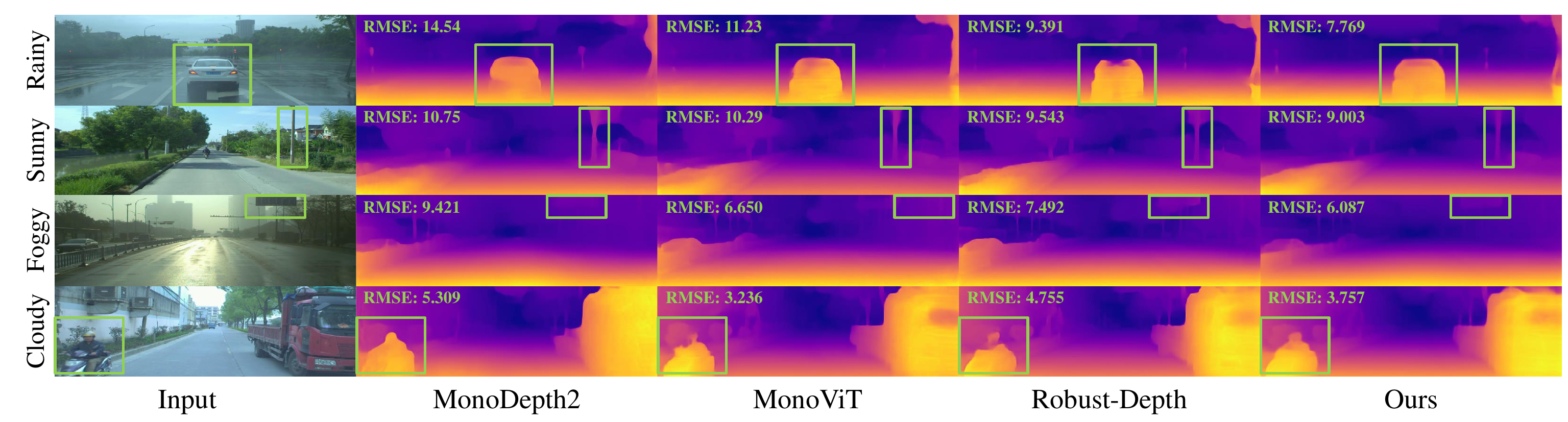}
    \vspace{-2em}
    \caption{\textbf{Qualitive results on DrivingStereo dataset.} Our method can recover detailed structures under different weather conditions in real scenes.}
    \label{fig:driving}
    \vspace{-0.5em}
\end{figure*}

\subsection{Evaluations}
In this section, we compare our models with the state-of-the-art methods.
In all tables, EC-Depth* represents the model of the first stage and EC-Depth represents the model of the second stage.

\textbf{Evaluation on KITTI~\cite{Geiger2013IJRR} and KITTI-C~\cite{kong2024robodepth}. } 
As shown in \Cref{table:results}, we compare our method with existing state-of-the-art methods on KITTI (standard) and KITTI-C (challenging) benchmarks.
Our first-stage model EC-Depth* outperforms other methods by a significant margin on the challenging benchmark while maintaining high performance on the standard benchmark, demonstrating the effectiveness of consistency regularization.
Additionally, the distilled model EC-Depth exhibits further performance improvement, surpassing our baseline MonoViT by 31.1\% in AbsREL on the KITTI-C benchmark, which proves the effectiveness of the proposed framework.
To vividly show the superiority of our method, we visualize the qualitative results in~\Cref{fig:kittic}. From the regions highlighted in green boxes, it is evident that our method can preserve detailed structures while other methods produce artifacts.

\begin{table}[t]
\caption{\textbf{Zero-shot evaluation on the DrivingStereo dataset.}}
\vspace{-0.5em}
\label{table:drivingstereo} 
\resizebox{\linewidth}{!}{
\begin{tabular}{c|ccc|ccc|ccc|ccc}
\hline
\multirow{2}{*}{Method} & \multicolumn{3}{c|}{\textbf{Cloudy}}                      & \multicolumn{3}{c|}{\textbf{Rainy}}                       & \multicolumn{3}{c|}{\textbf{Sunny}}                       & \multicolumn{3}{c}{\textbf{Foggy}}                        \\
                        & \cellcolor{low}AbsRel         & \cellcolor{low}SqRel          & \cellcolor{low}RMSE           & \cellcolor{low}AbsRel         & \cellcolor{low}SqRel          & \cellcolor{low}RMSE           & \cellcolor{low}AbsRel         & \cellcolor{low}SqRel          & \cellcolor{low}RMSE           & \cellcolor{low}AbsRel         & \cellcolor{low}SqRel          & \cellcolor{low}RMSE           \\ \hline
MonoDepth2\cite{godard2019digging}              & 0.170          & 2.211          & 8.453          & 0.245          & 3.641          & 12.282         & 0.177          & 2.103          & 8.209          & 0.143          & 1.952          & 9.817          \\
HR-Depth\cite{lyu2021hr}                & 0.173          & 2.424          & 8.592          & 0.267          & 4.270          & 12.750         & 0.173          & 1.910          & 7.924          & 0.154          & 2.112          & 10.116         \\
CADepth\cite{yan2021channel}                 & 0.161          & 2.086          & 8.167          & 0.226          & 3.338          & 11.828         & 0.164          & 1.838          & 7.890          & 0.141          & 1.778          & 9.448          \\
DIFFNet\cite{zhou_diffnet}                 & 0.154          & 1.839          & 7.679          & 0.197          & 2.669          & 10.682         & 0.162          & 1.755          & 7.657          & 0.125          & 1.560          & 8.724          \\
MonoViT\cite{zhao2022monovit}                 & \textbf{0.141} & 1.626          & 7.550          & 0.175          & 2.138          & 9.616          & \textbf{0.150} & 1.615          & 7.657          & 0.109          & 1.206          & 7.758          \\ 
Robust-Depth\cite{saunders2023self}            & 0.148          & 1.781          & 7.472          & 0.167          & 2.019          & 9.157          & 0.152          & 1.574          & \underline{7.293}          & \textbf{0.105} & 1.135          & 7.276          \\
\hline
EC-Depth*                & 0.149          & \underline{1.622}          & \underline{7.365}          & \textbf{0.162} & \textbf{1.723} & \textbf{8.478} & 0.153          & \underline{1.492}          & 7.317          & \underline{0.109}          & \underline{1.107}          & \underline{7.230}          \\
EC-Depth               & \underline{0.147}          & \textbf{1.561} & \textbf{7.301} & \textbf{0.162} & \underline{1.746}          & \underline{8.538}          & \underline{0.151}          & \textbf{1.436} & \textbf{7.213} & \textbf{0.105} & \textbf{1.061} & \textbf{7.121} \\ \hline
\end{tabular}
}
\vspace{-0.5em}
\end{table}

\begin{table}[t]
\caption{\textbf{Zero-shot evaluation on the NuScenes-Night dataset.}}
\vspace{-0.5em}
\label{table:nuscenes} 
\centering
\resizebox{0.8\linewidth}{!}{
\begin{tabularx}{\linewidth}{c|*{4}{C}|*{3}{C}}
\hline
Method      & \cellcolor{low}AbsRel & \cellcolor{low}SqRel & \cellcolor{low}RMSE & \multicolumn{1}{c|}{\cellcolor{low}RMSL} & \cellcolor{high}$ a_1 $ & \cellcolor{high}$ a_2 $ & \cellcolor{high}$ a_3 $             \\ \hline
MonoDepth2\cite{godard2019digging}    & 0.377                         & 4.689 & 11.625 & 0.506     & 0.394 & 0.683 & 0.826 \\
DIFFNet\cite{zhou_diffnet}      & 0.340                         & 3.912 & 10.608 & 0.448     & 0.448 & 0.731 & 0.865 \\
MonoViT\cite{zhao2022monovit}      & 0.304                         & 3.436 & 10.131 & 0.421     & 0.499 & 0.765 & 0.879 \\
Robust-Depth\cite{saunders2023self} & 0.286 & 3.795 & 9.220  & 0.379     & 0.594 & 0.822 & 0.912 \\ \hline
EC-Depth*     & \textbf{0.268}                         & \textbf{3.225} & \underline{8.990}  & \underline{0.356}     & \underline{0.604} & \underline{0.824} & \underline{0.920} \\
EC-Depth    & \underline{0.269}                         & \underline{3.347} & \textbf{8.902}  & \textbf{0.352}     & \textbf{0.613} & \textbf{0.831} & \textbf{0.924} \\ \hline
\end{tabularx}
}
\vspace{-1em}
\end{table}

\begin{table}[t]
\caption{\textbf{Ablation studies on different perturbations.} FP stands for feature-level perturbation. IP1 and IP2 stand for image-level perturbation.
}
\label{table:ablation_perb} 
\centering
\resizebox{0.9\linewidth}{!}{
\begin{tabularx}{1.1\linewidth}{c|*{3}{C}|*{4}{C}|*{3}{C}}
\hline
Benchmark                 & FP & IP1 & IP2 & \cellcolor{low}AbsRel & \cellcolor{low}SqRel & \cellcolor{low}RMSE & \multicolumn{1}{c|}{\cellcolor{low}RMSL} & \cellcolor{high}$ a_1 $ & \cellcolor{high}$ a_2 $ & \cellcolor{high}$ a_3 $ \\ \hline
                          &                            &                                &                                & 0.161                          & 1.285                         & 6.024                        & 0.245                             & 0.769                      & 0.916                      & 0.964                      \\
                          & \checkmark                          &                                &                                & 0.158                          & 1.275                         & 5.850                        & 0.240                             & 0.777                      & 0.920                      & 0.966                      \\
                          & \checkmark                          & \checkmark                              &                                & 0.116                          & 0.880                         & 4.817                        & 0.190                             & 0.867                      & \textbf{0.958}             & \textbf{0.982}             \\
\multirow{-4}{*}{KITTI-C} & \checkmark                          & \checkmark                              & \checkmark                              & \textbf{0.115}                 & \textbf{0.841}                & \textbf{4.749}               & \textbf{0.189}                    & \textbf{0.869}             & \textbf{0.958}             & \textbf{0.982}             \\ \hline
                          &                            &                                &                                & 0.100                          & 0.740                         & 4.427                        & 0.175                             & 0.900                      & 0.966                      & \textbf{0.984}                      \\
                          & \checkmark                          &                                &                                & \textbf{0.099}                 & 0.732                         & \textbf{4.366}               & \textbf{0.174}                    & \textbf{0.901}             & \textbf{0.967}             & \textbf{0.984}                      \\
                          & \checkmark                          & \checkmark                              &                                & 0.101                          & 0.719                         & 4.388                        & \textbf{0.174}                    & 0.894                      & 0.966                      & \textbf{0.984}                      \\
\multirow{-4}{*}{KITTI}   & \checkmark                          & \checkmark                              & \checkmark                              & 0.100                          & \textbf{0.708}                & 4.367                        & 0.175                             & 0.896                      & 0.966                      & \textbf{0.984}                      \\ \hline
\end{tabularx}
}
\vspace{-0.5em}
\end{table}

\textbf{Zero-shot evaluation on DrivingStereo~\cite{yang2019drivingstereo}. } 
To further demonstrate the robustness of our model, we do zero-shot evaluation on out-of-distribution datasets.
As shown in~\Cref{table:drivingstereo}, our method achieves state-of-the-art performance in four weather domains of DrivingStereo dataset.
It is worth noting that our model does not involve any rainy scenes during training, but it still outperforms Robust-Depth~\cite{saunders2023self}, which uses a physics-based renderer~\cite{tremblay2021rain} to simulate rainy day scenes for training.
~\Cref{fig:driving} displays the qualitative results of our method on DrivingStereo. It can be observed that our method can recover reasonable structures, laying a foundation for its deployment in autonomous driving.

\textbf{Zero-shot evaluation on NuScenes-Night~\cite{caesar2020nuscenes}. } 
Nighttime scenes pose another common challenge in autonomous driving due to their low visibility.
We conduct zero-shot testing on the NuScenes-Night dataset to validate the robustness of our method in nighttime scenes.
As shown in~\Cref{table:nuscenes}, experimental results demonstrate that our method surpasses previous state-of-the-art methods by a large margin, highlighting the superiority of our approach.

\subsection{Ablation Study}
In this section, we conduct detailed ablation studies on KITTI and KITTI-C benchmarks to demonstrate the effectiveness of the proposed components.
\textbf{Effectiveness of different perturbations.}
In \Cref{table:ablation_perb}, we demonstrate the effectiveness of different perturbations.
Firstly, perturbations at the feature level not only improve the depth prediction in challenging scenarios, but also exhibit an improvement in performance under standard conditions, demonstrating the complementary nature of feature-level perturbations and image-level perturbations. 
The last two rows compare the performance of using different numbers of strongly perturbed samples. 
Even with a single strong perturbation applied to the image, there is a significant improvement in the performance on the KITTI-C benchmark, with 28\% and 32\% decreases in AbsRel and RMSE, respectively. 
As we increase the number of strongly perturbed images, the performance on both KITTI and KITTI-C slightly improves.
Therefore, we choose two strongly perturbed images together with the weakly perturbed image to construct the image triplet.
The experimental results indicate that weak-to-strong perturbations can effectively explore the perturbation space and learn various depth priors.

\begin{table}[t]
\caption{\textbf{Ablation studies on consistency regularization.} PE stands for the modified photometric loss introduced in~\cite{saunders2023self}. CR means our consistency regularization. }
\label{table:ablation_regularization}
\centering
\resizebox{0.9\linewidth}{!}{
\begin{tabularx}{1.1\linewidth}{c|*{2}{C}|*{4}{C}|*{3}{C}}
\hline
Benchmark                 & PE &  CR & \cellcolor{low}AbsRel & \cellcolor{low}SqRel & \cellcolor{low}RMSE & \multicolumn{1}{c|}{\cellcolor{low}RMSL} & \cellcolor{high}$ a_1 $ & \cellcolor{high}$ a_2 $ & \cellcolor{high}$ a_3 $ \\ \hline
                        &                        &                                & 0.161                          & 1.285                         & 6.024                        & 0.245                             & 0.769                      & 0.916                      & 0.964                      \\
                          & \checkmark                                  &                            & 0.118                          & 0.966                          & 4.907                        & 0.195                             & \textbf{0.874}                      & 0.957                      & 0.980                      \\
\multirow{-3}{*}{KITTI-C} &                                 & \checkmark                          & \textbf{0.115}                 & \textbf{0.841}                & \textbf{4.749}               & \textbf{0.189}                    & 0.869             & \textbf{0.958}             & \textbf{0.982}             \\ \hline
                          &                               &                                & 0.100                          & 0.740                         & 4.427                        & 0.175                             & \textbf{0.900}                      & 0.966                      & \textbf{0.984}                      \\
                          & \checkmark                                          &                            & 0.102                          & 0.799                         & 4.494                        & 0.179                             & 0.898                      & 0.965                      & 0.983                      \\
\multirow{-3}{*}{KITTI}   &                                      & \checkmark                          & \textbf{0.100}                 & \textbf{0.708}                & \textbf{4.367}               & \textbf{0.175}                    & 0.896                      & \textbf{0.966}                      & \textbf{0.984}             \\ \hline
\end{tabularx}
}
\vspace{-1em}
\end{table}

\begin{table}[t]
\begin{minipage}[t]{0.48\linewidth}
% \begin{table}
\centering
\caption{\textbf{Ablation studies on depth consistency loss.} $L_{SI}$ stands for the scale-invariant loss~\cite{eigen2014depth}, and $L_{c}$ stands for the proposed depth consistency loss.
}
\label{table:ablation_lc}
\resizebox{\linewidth}{!}{
\begin{tabularx}{1.3\linewidth}{c|*{1}{C}|*{4}{C}}
\hline
Test                      & Loss & \cellcolor{low}AbsRel & \cellcolor{low}SqRel & \cellcolor{low}RMSE & \multicolumn{1}{c}{\cellcolor{low}RMSL} \\ \hline
                          & $L_{SI}$            & 0.123                          & 0.938                         & 4.948                        & 0.197                             \\
\multirow{-2}{*}{KITTI-C} & $L_c$             & \textbf{0.115}                 & \textbf{0.841}                & \textbf{4.749}               & \textbf{0.189}                    \\ \hline
                          & $L_{SI}$            & \textbf{0.099}                          & 0.734                         & \textbf{4.362}                        & 0.176                             \\
\multirow{-2}{*}{KITTI}   & $L_c$             & 0.100                 & \textbf{0.708}                & \textbf{4.367}               & 0.175                    \\ \hline
\end{tabularx}
}
% \end{table}
\end{minipage}
\hspace{1em}
\begin{minipage}[t]{0.48\linewidth}
% \begin{table}
\centering
\caption{\textbf{Ablation studies on Mean Teacher.}
IST stands for iterative self-training, and MT stands for the Mean Teacher paradigm.}
\label{table:ablation_meanteacher}
\resizebox{\linewidth}{!}{
\begin{tabularx}{1.3\linewidth}{c|c|*{4}{C}}
\hline
Test                      & Framework     & \cellcolor{low}AbsRel & \cellcolor{low}SqRel & \cellcolor{low}RMSE & \multicolumn{1}{c}{\cellcolor{low}RMSL} \\ \hline
                          & IST    & 0.112                          & 0.823                         & 4.674                        & 0.186                             \\
\multirow{-2}{*}{KITTI-C} & MT & \textbf{0.111}                 & \textbf{0.807}                & \textbf{4.651}               & \textbf{0.185}                    \\ \hline
                          & IST    & 0.100                          & 0.699                         & 4.327                        & 0.173                             \\
\multirow{-2}{*}{KITTI}   & MT & \textbf{0.100}                 & \textbf{0.689}                & \textbf{4.315}               & \textbf{0.173}                    \\ \hline
\end{tabularx}
}
% \end{table}
\end{minipage}
\end{table}

\textbf{Efftectiveness of consistency regularization.}
We compare the proposed consistency regularization with another solution introduced by~\cite{saunders2023self} in \Cref{table:ablation_regularization}.
The first row in the table stands for our baseline.
While the modified photometric loss are introduced in~\cite{saunders2023self} to improve the depth prediction in challenging scenarios, it results in a slight performance degradation on the KITTI dataset. 
We speculate that since the photometric loss essentially treats the perturbed image as a new sample, the model needs to do trade-off between these two domains.
However, exploiting consistency regularization to constrain the depth consistency under perturbations provides a more direct way to propagate supervision to challenging scenarios.
It not only upholds its performance in the standard domain but also enhances the robustness under diverse perturbations.

\begin{table}[t]
\caption{\textbf{Ablation studies on consistency-based pseudo-label filtering.} GC means geometric consistency-based filter. DC means depth consistency-based filter. 
% % In stage 2, we use the student network for evaluation.
}
% \vspace{-1em}
\label{table:ablation_mask} 
\centering
\resizebox{0.9\linewidth}{!}{
\begin{tabularx}{1.1\linewidth}{c|*{2}{C}|*{4}{C}|*{3}{C}}
\hline
Benchmark                 & GC & DC & \cellcolor{low}AbsRel & \cellcolor{low}SqRel & \cellcolor{low}RMSE & \multicolumn{1}{c|}{\cellcolor{low}RMSL} & \cellcolor{high}$ a_1 $ & \cellcolor{high}$ a_2 $ & \cellcolor{high}$ a_3 $ \\ \hline
                          &                                     &                                     & 0.115                          & 0.841                         & 4.749                        & 0.189                             & 0.869                      & 0.958                      & 0.982                      \\
                          & \checkmark                                   &                                     & 0.113                          & 0.841                         & 4.637               & 0.186                             & 0.874                      & \textbf{0.960}             & \textbf{0.983}             \\
                          &                                     & \checkmark                                   & 0.112                          & 0.825                         & 4.641                        & \textbf{0.185}                    & \textbf{0.875}             & \textbf{0.960}             & \textbf{0.983}             \\
\multirow{-4}{*}{KITTI-C} & \checkmark                                   & \checkmark                                   & \textbf{0.111}                 & \textbf{0.807}                & \textbf{4.651}               & \textbf{0.185}                    & 0.874             & \textbf{0.960}             & \textbf{0.983}             \\ \hline
                          &                                     &                                     & \textbf{0.100}                          & 0.708                         & 4.367                        & 0.175                             & 0.896                      & 0.966                      & 0.984                      \\
                          & \checkmark                                   &                                     & 0.101                 & 0.709                         & 4.310               & \textbf{0.173}                    & 0.896             & \textbf{0.967}             & \textbf{0.985}             \\
                          &                                     & \checkmark                                   & \textbf{0.100}                          & 0.699                         & \textbf{4.305}               & \textbf{0.173}                    & \textbf{0.897}                      & \textbf{0.967}             & \textbf{0.985}             \\
\multirow{-4}{*}{KITTI}   & \checkmark                                   & \checkmark                                   & \textbf{0.100}                          & \textbf{0.689}                & 4.315                        & \textbf{0.173}                    & 0.896                      & \textbf{0.967}             & \textbf{0.985}             \\ \hline
\end{tabularx}
}
\vspace{-1em}
\end{table}

\textbf{Effectiveness of perturbation-invariant depth consistency loss.}
Here, we investigate the impact of different losses on consistency regularization. 
As shown in \Cref{table:ablation_lc}, when applying the proposed perturbation-invariant depth consistency loss in the first-stage training, it outperforms the model using the scale-invariant loss~\cite{eigen2014depth} by a large margin on KITTI-C benchmark. 
This compellingly demonstrates the superiority of the proposed depth consistency loss.

\textbf{Effectiveness of the Mean Teacher paradigm.}
Iterative self-training updates pseudo-labels offline at each training round and iteratively optimizes the network using the updated pseudo-labels. 
As shown in \Cref{table:ablation_meanteacher}, Mean Teacher yields higher accuracy and robustness than iterative self-training.
This suggests that Mean Teacher paradigm can generate more accurate pseudo-labels through convenient information transfer between the teacher and the student.

\textbf{Effectiveness of consistency-based pseudo-label filtering strategy.}
In \Cref{table:ablation_mask}, we investigate the effect of the consistency-based pseudo-label filtering strategy. Adopting the geometric consistency-based filter alone or the depth consistency-based filter alone both contributes to improved performance and robustness. 
This indicates that both filtering strategies can effectively enhance the quality of pseudo-labels. 
Furthermore, when combining both filters, the performance is further improved, highlighting the compatibility of the two strategies.

\section{Conclusion}
In this paper, we propose EC-Depth, which explores the consistency of self-supervised monocular depth estimation to improve the accuracy and robustness of the model especially in challenging scenarios. 
The proposed EC-Depth includes two training stages.
In the first stage, we leverage consistency regularization to construct effective supervision for depth predictions of challenging images.
In the second stage, we distill the model of the first stage by Mean Teacher and devise a consistency-based pseudo-label filtering strategy.
Quantitative and qualitative results demonstrate our technical contribution and the effectiveness of the proposed architectural design.
The proposed EC-Depth is agnostic to the design of a specific network branch, which allows it to be easily transferred to any self-supervised monocular depth estimation method.

\clearpage  % TODO REVIEW/FINAL: This \clearpage needs to be removed from both review and camera-ready versions.

\bibliographystyle{splncs04}
\bibliography{main}
\end{document}